\documentclass[11pt,a4paper]{article}
\usepackage[hyperref]{acl2019}
\usepackage{times}
\usepackage{latexsym}

\usepackage{url}

\usepackage{graphicx}
\usepackage{xcolor}
\usepackage{balance}
\usepackage{tabularx}

\aclfinalcopy

\begin{document}
\author{Brendan Bena\\
Drury University\\
900 N. Benton Ave.\\
Springfield, Missouri 65109
\And
Jugal Kalita\\
UC-Colorado Springs\\
1420 Austin Bluffs Pkwy.\\
Colorado Springs, Colorado 80918
}

\title{Introducing Aspects of Creativity in Automatic Poetry Generation}

\maketitle
\begin{abstract}
Poetry Generation involves teaching systems to automatically generate text that resembles poetic work. A deep learning system can learn to generate poetry on its own by training on a corpus of poems and modeling the particular style of language. In this paper, we propose taking an approach that fine-tunes GPT-2, a pre-trained language model, to our downstream task of poetry generation. We extend prior work on poetry generation by introducing creative elements. Specifically, we generate poems that express emotion and elicit the same in readers, and poems that use the language of dreams---called dream poetry. We are able to produce poems that correctly elicit the emotions of \textit{sadness} and \textit{joy} 87.5 and 85 percent, respectively, of the time. We produce dreamlike poetry by training on a corpus of texts that describe dreams. Poems from this model are shown to capture elements of dream poetry with scores of no less than 3.2 on the Likert scale. We perform crowdsourced human-evaluation for all our poems.  We also make use of the Coh-Metrix tool, outlining  metrics we use to gauge the quality of text generated.
\end{abstract}
\section{Introduction}
Many natural language processing tasks require the generation of human-like language. Some tasks, such as image and video captioning and automatic weather and sports reporting, convert non-textual data to text.
Some others, such as summarization and machine translation, convert one text to another.  There are additional tasks  that aim  to produce text, given a topic or a few keywords such as story generation, joke generation, and poetry generation, among others.

Poetry generation produces creative content, and delivers the content  in an aesthetically pleasing manner, usually following a specific structure. Thus, in addition to generating text as if in a story, the lines produced usually have a certain length, quite frequently there is a rhyming scheme as well as rhythm, and organization into structures such as couplets, quatrains, quintets, and stanzas. Among other tools, creativity comes from unusual usage of words through effects such as alliteration, assonance, and elision; use of metaphors, symbolism, and other linguistic devices; licensing of underlying imagery with expressed feelings, sentiments, and emotions.

Work in natural language generation can be traced to pioneering rule-based simulations of chatbots such as the  ``psychotherapist" Eliza \cite{Weizenbaum1966} and  paranoid schizophrenia-suffering PARRY \cite{Colby1981}. Surveys such as \cite{Hovy1990,Reiter2000,Gatt2018,Santhanam2019} have described the progress in natural language generation over 50 years. Of late, the use of deep learning has produced enviable progress in natural language generation, especially in topics such as machine translation \cite{Bahdanau2014,Wu2016}, image captioning \cite{Mao2014} and dialogue generation \cite{Li2016}. 

This paper discusses the automatic generation of natural-sounding poems that are creative. Creativity comes in many hues, and we experiment with a few established ways of creative expression in poetry generation. First, we generate poetry that can potentially evoke a response from the readers or hearers in terms of emotions and feelings they generate. Additionally, we choose the idea of mimicking the language of dreams as another form of creative expression due to its longstanding history in poetry. Dream poetry  dates back to medieval times where famous fourteenth century authors, like Chaucer, experimented using dreams as the structure for an image or picture they wished to paint with a poem \cite{medievaldreams}. A dream poem is said to be characterized by the `I' of the poem and its substance of a dream or a vision included \cite{medieval}. To the best of our knowledge, prior work on poetry generation, whether using deep learning or not, has not explored the incorporation of emotion-eliciting phraseology or elements of creativity such as dream  poetry.

Our research provides the following contributions:
\vspace{-5pt}
\begin{itemize}
\setlength{\itemsep}{-5pt}
  \item generating grammatical, coherent, and flowing poetry using the powerful and versatile GPT-2 architecture, 
  \item successfully generating poetry that elicits certain emotions in readers, and
  \item generating poems that follow time-honored tradition of dream-like language usage and imagery.
\end{itemize}

This paper is organized as follows. Section 2 presents related work. Section 3 discusses our approach to creative text generation including pre-processing steps, architecture used, and approaches to training. Section 4 discusses our experiments and results. Finally, we present  evaluation of our research in Section 5, followed by conclusions and future work in Section 6.
\section{Related Work}
Early methods for poetry generation made use of template-oriented and rule-based techniques. These approaches often required a large amount of feature picking and knowledge of syntactic and semantic rules in a language \cite{oliveira2009automatic,oliveira2012poetryme}. Other methods treated  poetry generation as special cases of machine translation or summarization tasks \cite{yan,he}. We believe that forcing a model to adhere to specific rules or templates, or summarizing or translating a given text to generate new poetry is unlikely to lead to the artistically expressive quality we seek to generate. 

More recently, deep learning methods have become prevalent in natural language generation, including poetry generation. \citeauthor{Zhang2014}  (\citeyear{Zhang2014}) for instance, used Convolutional (CNN) and Recurrent  Neural Networks (RNN) to generate Chinese Poetry. RNNs allow for short-term memory of the language to be maintained by inputting the generated output of a network  cell back into itself, essentially building context. \citeauthor{Ghazvininejad2017} (\citeyear{Ghazvininejad2017}) used Long Short-Term Memory (LSTM) units, which are advanced gated versions of RNNs,   to the task of poetry generation. \citeauthor{Wei2018} (\citeyear{Wei2018}) attempted to address the style issue by training the networks using particular poets and controlling for style in Chinese poetry. They found that with enough training data, adequate results could be achieved. Problems related to poetic structure were addressed by \citeauthor{Hopkins2017} (\citeyear{Hopkins2017}). They generated rhythmic poetry by training the network on only a single type of poetry to ensure produced poems adhered to a single rhythmic structure. It was found in human evaluations that while the poems produced were rated to be of lower quality than human produced poems, they were indistinguishable from human produced poems. \citeauthor{Lau2018} (\citeyear{Lau2018}) took the LSTM approach one step further with the \textit{Deepspeare} model by employing an attention mechanism to model interactions among generated words. They also use three neural networks, one for rhythm, one for rhyming and another for word choice in their quest to generate Shakespeare-like sonnets. 

\citeauthor{Vaswani2017} (\citeyear{Vaswani2017}) developed a deep neural architecture called the Transformer that did away with any sort of need for recurrence. The Transformer also employed an elaborate attention mechanism that has been  shown to  be useful in natural language tasks. \citeauthor{Radford2019} (\citeyear{Radford2019}) used this architecture in their Generative Pretrained Transformer 2 (GPT-2) model. GPT-2 is capable of many downstream tasks like text generation but to our knowledge, research has not been published using the GPT-2 model specifically for poetry generation. 

On a slightly different but related note,  natural language generation influenced by multi-modal input was attempted by \citeauthor{Vechtomova2018} (\citeyear{Vechtomova2018}) to generate song lyrics in the style of specific artists by fusing outputs coming from lyrical inputs processed by an RNN and audio clips processed by a CNN. Text generation has also been influenced, in a cross domain manner, through images. The works of \citeauthor{Liu:2018} (\citeyear{Liu:2018}) have shown that coupled visual-poetic embeddings can be used to pick out poetic clues in images, which in turn can be used to inspire the generated text. Though influenced natural language generation in and of itself is not a novel idea, we feel our attempt to style text with the intent of eliciting particular emotions provides a creative way to explore this subtask.
\section{Approach}
Our goal is to successfully demonstrate the introduction of creative flair in automatic poetry generation in two exemplar ways: explicit show of emotion and the use of language that is predominantly first person with dream-like imagery. To enable the expression of emotion in generated poems, our work involves a preliminary step of scoring a corpus of downloaded poems for emotion to produce subsets of poems that express one of eight different identified emotions. This step is followed by the actual generation of poems by fine-tuning the pre-trained GPT-2 natural language model. We train eight separate models for eight different emotions, each on a sub-corpus predominantly demonstrating a particular emotion. To generate poems that use dream-like language, we create a text corpus composed of  a large number of dream transcriptions created in first person by actual viewers of dreams. In this case,  we  apply transfer learning by fine-tuning the pre-trained GPT-2 on the dream corpus, followed by training again on poetry. We evaluate the generated poems using automated techniques as well as humans.

\subsection{Poem Emotion Scoring}

\begin{figure}[h!]
    \centering
    \includegraphics[width=0.75\linewidth]{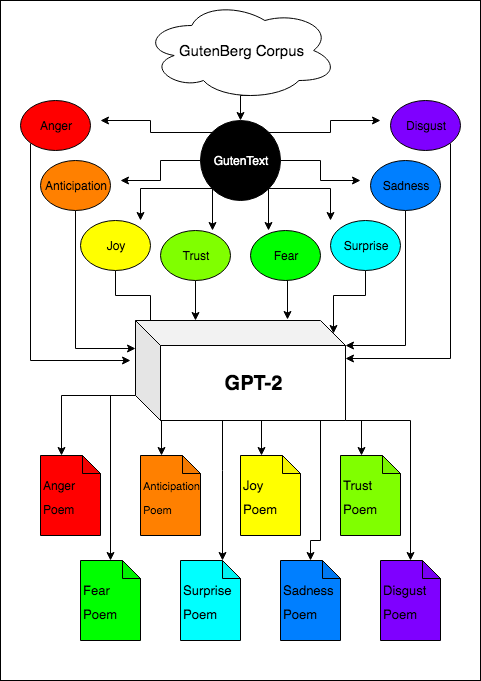}
    \caption{A high-level overview of our project implementation for emotion eliciting poetry} 
    \label{fig:overview}
\end{figure}

A high-level overview of the emotion elicitation portion of our project is shown in Figure \ref{fig:overview}. To create a corpus of poems based on  the emotions they elicit, we make use of the EmoLex dictionary \cite{Mohammad}. EmoLex is a word-level emotion lexicon that associates English words with the eight different emotion categories we wish to explore. Each poem (or book of poems) in our dataset is given a score that is the total of the associated emotion scores in EmoLex for each word. The maximum emotion word score is taken and the poem is labeled under that emotion category. We create eight such datasets, one corresponding to each emotion category supported by EmoLex. This approach allows us to   to train multiple models on our split dataset.

Currently, the emotions of \textit{joy, anticipation, trust, anger}, and \textit{sadness} represent a large portion of our data while the emotions of \textit{surprise, disgust,} and \textit{fear} are severely underrepresented. Table \ref{tab:models} shows key differences in models including the number of tokens in the text and the final average loss during training.

\subsection{GPT Architecture}

To create a model for poetic language, we propose finetuning OpenAI's GPT-2 architecture. GPT-2 is a Transformer-based model that was trained simply to predict the next word in a 40GB text corpus \cite{Radford2019}. This 40GB dataset, \textit{WebText}, was scraped from the internet with certain heuristics that aimed to gather only quality text (i.e. only outbound Reddit links from posts with a karma rating of 3 stars or better). By training on such a large, all-encompassing corpus of text, the architecture has proven to model the English language well and has obtained state-of-the-art results on  downstream text-based tasks such as machine translation, question answering, and summarization. We leverage GPT-2's pre-trained knowledge of language for our downstream task of peotry generation.

\begin{figure}[h!]
    \centering
    \includegraphics[width=.65\linewidth]{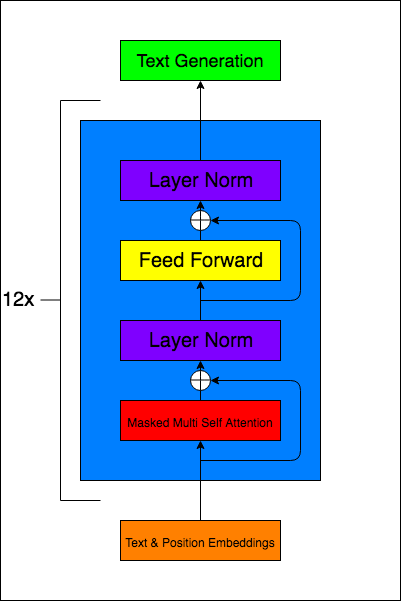}
    \caption{GPT Architecture. Adapted from \cite{gpt,Radford2019}}
    \label{fig:GPT}
\end{figure}

GPT-2  is the successor of OpenAI's first Transformer-based architecture, GPT \cite{gpt}, with a few changes to the structure. The medium version of GPT-2 we use contains 345M parameters and is a 24 layer, decoder-only Transformer architecture. GPT-2 moves  layer normalization  to the input of each sub-block, adds another layer normalization after the final self-attention block and  increases context size from 512 to 1024 tokens. This architecture allows for long term dependencies to be captured better in language modeling. GPT-2's attention mechanism is referred to as a masked multi self-attention head. This technique allows for a relationship to be modeled for all words in an input sequence. Words that have multiple meanings can then be represented based on the context they appear in. Higher attention scores from surrounding words relate to a larger contribution to the representation of a word. GPT-2 makes use of byte-pair encoding (BPE) like its predecessor GPT but on UTF-8 byte sequences \cite{sennrich2015neural}. GPT-2's encoding is somewhere in between character level and word level. The model also prevents different versions of common words from being duplicated (i.e. \textit{fate!}, \textit{fate?}, and \textit{fate} would not be joined). This technique improves the quality of the final byte segmentation. GPT-2's encoding rids the need for pre-processing or tokenization of data and is able to assign a probability to any Unicode string. 

\begin{table}[]
    \begin{center}
        \begin{tabular}{ | p{0.5in} | c | c | p{0.5in} | } 
            \hline
             \bf{Data} & Model Size &  \# of Tokens & Final Loss \\ 
            \hline
            {anger} & 345M & 1,292,457 & 0.27 \\ 
            \hline
            {antici-pation} & 345M & 2,314,637 & 1.30 \\ 
            \hline
            {joy} & 345M & 11,668,792 & 3.19 \\ 
            \hline
            {sadness} & 345M & 2,090,915 & 1.03 \\ 
            \hline
            {trust} & 345M & 16,667,178 & 3.39 \\ 
            \hline
        \end{tabular}
    \end{center}
\caption{Comparison of 5 emotion models trained.}
\label{tab:models}
\end{table}

\subsection{Training for Creative Poem Generation}
The task-agnostic nature of GPT-2 allows us to take a fine-tuning approach to our downstream task of poetry generation. Our approach to generating poems that exhibit emotion as well as dream-like imagery involves training the pre-trained GPT-2 model. Our training protocol for the two cases are stated briefly below. 

\subsubsection{Generating Emotion Poems}
Poetry is a personal form of writing that expresses human feelings, and  \citeauthor{Mill1860} (\citeyear{Mill1860}) famously  said ``What is poetry, but thoughts and words in which emotion spontaneously embodies itself?" \citeauthor{Mill1833} (\citeyear{Mill1833}) also said "The object of poetry is confessedly to act upon the emotions''. Expressing emotions, with possible motive of eliciting the same emotions in readers, is a basic characteristic of poems. Our goal in this paper is to use artificial neural networks to generate poems that explicitly evoke certain specific emotions. 

To generate poems with emotional content, we have split our poetry data into sub-corpora, one sub-corpus for each emotion. We train the already pre-trained GPT-2 on a sub-corpus of poems that demonstrate a certain emotion. Pre-trained GPT-2 has a very strong foundational knowledge of English. We find that training it again on emotion-bearing poetry seems to enable it to generate high quality poetry, which is even able to use emotion-laden words for the correct form of elicitation. We also find that the poems we generate seem to exhibit proper punctuation as well as lines that have poem-appropriate length and sentences that are grammatically correct. In addition, the poems we generate seem to be quite readable and demonstrate high coherence. Detailed analyses are reported in the next section.  

\subsubsection{Generating Dream Poems}
Dream poems represent a style of poetry that was ``astonishingly" popular in the 14th through the 16th centuries \cite{Spearing1976,Windeatt2003} and are still popular \cite{Russo2003}. Such poems tell a story based on a dream or a number of dreams, dreamt by the narrator or by a character that the poet introduces. \citeauthor{Spearing1976} (\citeyear{Spearing1976}) claimed that dream poems are based on objective experience, but at the same time they are free of constraints of everyday possibilities. Such poems represent the outcome of a poetic process with many different influences, models, and analogues \cite{Windeatt2003}, but without going into such details,  our goal is to see if an ANN can produce poems which share characteristics with dream poems. 

To generate poems that demonstrate  first-person language with dream-like imagery, we take a similar approach. However, in this case, GPT-2 undergoes three separate training cycles. The first cycle is the pre-training that GPT-2 goes through before release to the public by OpenAI. Second, we train the pre-trained model on a corpus of first-person dream descriptions. Third, we train again on poems. Our hypothesis is that pre-training by OpenAI results in good basic knowledge of English; that training on the dream corpus endows the network with the knowledge of first-person imagery-based language; and that the last training cycle teaches the network language of poems. We demonstrate in the next section that we are not far off from our being successful in our hypothesis. 

\subsection{Text Generation and Sampling}

As stated by \citeauthor{Radford2019} (\citeyear{Radford2019}), the core approach of GPT-2 is language modeling. A language model can be thought of as a probability distribution over a sequence of words in the form:

\begin{equation}
    p(w_{1},...,w_{n}).
\end{equation}

Likewise, natural language tends to have a sequential order so it can be modeled in terms of the conditional probability of a word given the words preceding it \cite{bengio2003neural}:

\begin{equation}
    p(w_{n}|w_{1},...,w_{n-1}).
\end{equation}

We make use of this probabilistic style of language modeling by sampling from the distribution in a semi-random fashion. Just as the GPT-2 paper does for its text generation, we make use of Top K sampling, limiting the possible guesses of words to 40. In addition to Top K, we make use of a temperature constant of 0.75 which controls randomness in the distribution. A temperature closer to 0 correlates to less randomness while a temperature closer to 1 relates to more randomness. Finally, at the end of the generation process, we employ a simple text cleaning algorithm that allows poems to end more naturally and rather than trail off as they do sometimes.
\section{Experiments and Results}

\subsection{Datasets and Resources}
In order to classify emotion-eliciting poems or books, we use the NRC Word-Emotion Association Lexicon (EmoLex) resource. EmoLex\footnote{\url{https://saifmohammad.com/WebPages/NRC-Emotion-Lexicon.htm}} was created by the National Research Council of Canada and includes 14,182 English words that are associated with different emotions and positive or negative sentiment \cite{Mohammad}. Words in EmoLex have been manually annotated via crowd-sourcing and emotions fall into one or more categories of eight basic emotions: \textit{joy, trust, fear, surprise, sadness, anticipation, anger, and disgust} \cite{principles}. We elect to use this simplified version of the Wheel of Emotions due to its parallels with the available EmoLex dataset.  This resource provides us with a way to fabricate a ground truth in the types of emotion-infused texts we wish to use for training data.

\begin{figure}[h!]
    \centering
    \includegraphics[width=0.50\linewidth]{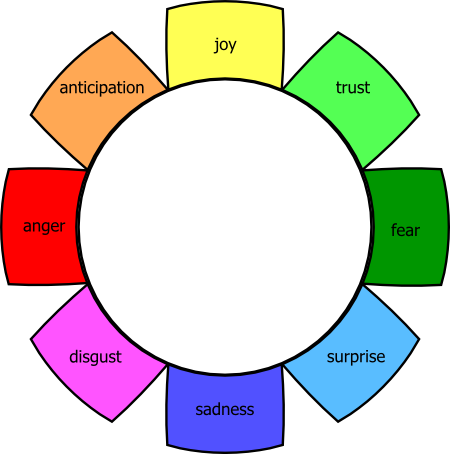}
    \caption{American pyschologist Robert Plutchik's Wheel of Emotions} 
    \label{fig:EmoLex}
\end{figure}

To handle the training and generation portions of the project, we draw data from the Project Gutenberg website\footnote{\url{https://www.gutenberg.org/}}. Project Gutenberg is a massive online database containing over 59,000 eBooks. We limit this corpus to a smaller subcorpus using an adaptation of the GutenTag tool \cite{gutentag}. This tool allows us to place constraints on the amount of literature we choose to use in our work. Our final dataset includes approximately three million lines of poetic text from the Gutenberg database and is further divided by poem/book into our eight emotion categories.

We attempt to create dream poetry by making use of the \textit{DreamBank} dataset. The \textit{DreamBank} was created by Schneider \& Domhoff at UC-Santa Cruz\footnote{\url{https://www.dreambank.net/}}. The dataset contains a collection of over 20,000 dreams from users age 7 to 74. We scraped this dataset from the website assuring that dreams collected were recorded only in English. The \textit{DreamBank} allows us to attempt transfer learning by finetuning on the dream dataset first, then further finetuning on our poetry dataset.

Initially, we retrained 6 GPT-2 based models. Default training parameters were used each of the 5 different emotion datasets and our dream dataset. All were trained for 12,000 steps (except for our dream model which was trained for 12k steps on dreams and on poetry) with a learning rate of 0.0001. When generating text, we do not input context: we allow the model to write the poem entirely through the sampling of conditional probability from the language it has modeled.

Figures 4 through 8 give examples of 5 poems that we have hand-picked to illustrate the quality of poems generated. A cursory glance at the poems reveals the high quality of the text in terms of lexical choice, grammatical integrity, and semantic cohesion. We discuss how we quantitatively assess the poems below.

\begin{figure}
\begin{center}
\begin{tabular}{|c|}
\hline
Heard I a song of joy,\\
A song of happy sound,\\
Fills all the air I breathe,\\
To him I sing, to him\\
I sing the happy song.\\
All night long on the steep green grass\\
I ride and sing\\
\hline
\end{tabular}
\label{fig:joypoem}
\end{center}
\caption{A hand-picked, automatically generated poem from the joy model}
\end{figure}

\begin{figure}
\begin{center}
\begin{tabular}{|c|}
\hline
The other, who with one accord\\
Wrote my essay, in that he was dear\\
And good, and knew well, how we ought to treat\\
A man of such renown, and such love?\\
He's a good honest man, no doubt\\
\hline
\end{tabular}
\label{fig:trustpoem}
\end{center}
\caption{A hand-picked, automatically generated poem from the trust model}
\end{figure}

\begin{figure}
\begin{center}
\begin{tabular}{|c|}
\hline
We have reached the peak of the highest mountain \\ \makebox[2in]{}in the world\\
The mountain of dreams.\\
This is the view\\
Across the valley,\\
One hour's journey back,\\
We crossed it on the way between\\
A band of beautiful young women.\\
There was\\
\hline
\end{tabular}
\label{fig:anticpoem}
\end{center}
\caption{A hand-picked, automatically generated poem from the anticipation model}
\end{figure}

\begin{figure}
\begin{center}
\begin{tabular}{|c|}
\hline
A long trail of falling mist\\
Had made its way here, and now\\
Aerily it seemed, as if to drown\\
The discordant thunder clang.\\
It seemed to drown the music of the rain;\\
In this lost place of sorrow\\
Far off\\
\hline
\end{tabular}
\label{fig:sadpoem}
\end{center}
\caption{A hand-picked, automatically generated poem from the sadness model}
\end{figure}

\begin{figure}
\begin{center}
\begin{tabular}{|c|}
\hline
Amidst the chaos throng'd, with angry voices each\\
His rival's mockery; loud their scorn was fill'd;\\
So fierce their rage, and in their eager power\\
Met on the walls of Troy, were fill'd with dismay.\\
\hline
\end{tabular}
\label{fig:angerpoem}
\end{center}
\caption{A hand-picked, automatically generated poem from the anger model}
\end{figure}

\begin{figure}
\begin{center}
\begin{tabular}{|c|}
\hline
A thousand stars at once,\\
An hundred thousand stars!\\
The sun was low,\\
And the stars were bright,\\
My heart would do the same.\\
A thousand stars at once,\\
A hundred thousand stars!\\
The night had begun,\\
And the stars were all the same.\\
When I came back from the dead,\\
I saw the stars\\
\hline
\end{tabular}
\label{fig:dreampoem1}
\end{center}
\caption{A hand-picked, automatically generated poem from the dream model}
\end{figure}

\begin{figure}
\begin{center}
\begin{tabular}{|c|}
\hline
For she was mine.\\
I was the only one\\
She had,\\
And a thousand other friends,\\
And a hundred more\\
She held me dear.\\
Her eyes were clear, her cheeks were bright,\\
Her heart was like a rose,\\
Her mouth was full of music,\\
Her lips were white\\
As snow,\\
And the music she sang\\
\hline
\end{tabular}
\label{fig:dreampoem2}
\end{center}
\caption{A hand-picked poem, automatically generated from the dream model}
\end{figure}

\section{Evaluation}
In the first crowd-sourced analysis of our emotion-eliciting poetry we presented
four poems from each category (of the five data-represented emotion categories)
to ten human reviewers with undergraduate level educational backgrounds. All
reviewers are native speakers of English. Poems presented were randomly selected
from the top 20 EmoLex scored poems out of a pool of 1,000 generated poems.
These reviewers were asked to rate each poem based on the emotions elicited
within them after reading. An emotion was deemed correctly
elicited if the associated Likert score was 4 or greater from the reviewer.
Table \ref{tab:emotion} illustrates the results from our evaluation. When taking
the average percentage of correct emotion-eliciting poems, the models of joy,
sadness, and anger produced the most promising results while the trust and
anticipation models were less than satisfactory. We believe this is because joy,
sadness and anger are basic or fundamental emotions compared to trust and
anticipation, which are more complex and difficult to explain. Although there
are many opinions among psychologists about what constitute basic emotions, joy,
sadness and anger, (especially the last two) seem to occur the most often in
proposals that demarcate a set of basic emotions \cite{Ortony1990}.

\begin{table}[]
\begin{center}
\begin{tabular}{ | c | c | c | c | c | c | } 
\hline
\bf{Emotion} & {Anger} & {Antic.} & {Joy} & {Sad.} & {Trust} \\ 
\hline
\bf{\%} & 65 & 40 & 85 & 87.5 & 32.5 \\ 
\hline
\end{tabular}
\end{center}
\caption{Average percentage of correctly elicited emotion across four poems in each category}
\label{tab:emotion}
\end{table}

To preserve consistency in our experiments, we evaluate our dream model poetry in a manner similar to our evaluation of the emotion poems. Four poems from the model were presented to the same ten judges and they were asked to assess the poems based on qualities of dream poetry. These poems were cherry picked from a pool of 1,000 generated poems. A dream poem is said to have the following qualities \cite{Windeatt2003,Spearing1976,Russo2003} among many other qualities. We believe these three are the least ambiguous and easiest to decipher for human evaluation. 
\vspace*{-5pt}
\begin{itemize}
\setlength{\itemsep}{-5pt}
  \item Quality 1: The poem is generally a first-person expression
  \item Quality 2: The poem's main substance is dream or vision like
  \item Quality 3: The poem recounts or foretells an experience or event 
\end{itemize}

Analysis of results show that machine generated poems are able to capture the first person perspective well, achieving between 4.5 and 5 average Likert scores. The poems often appear to retell a story or an event,  scoring between 3.7 and 4.2 average Likert scores. The nature of poetry and dream recounts that make up our data is often narrative, so this result stands to reason. However, Quality 2 scores of the poem substance containing a dream or vision are questionable. We suspect the Quality 2 score is lower due to the ambiguity in ascertaining dream text from regular text. Table \ref{tab:dream} highlights our results for the dream model.

\begin{table}[]
\begin{center}
\begin{tabular}{| c | c | c | c | c |}
\hline
\textbf{Poem} & 1 & 2 & 3 & 4 \\
\hline
Qual 1 & \textbf{5} & 4.9 & 4.8 & 4.5 \\
\hline
Qual 2 & 3.5 & \textbf{4.1} & 3.2 & 3.3\\
\hline
Qual 3 & 3.9 & \textbf{4.2} & 3.7 & 3.7\\
\hline
\end{tabular}
\end{center}
\caption{Average Likert score of users for each poem}
\label{tab:dream}
\end{table}

Currently, there exists no widely available standard for evaluating poetry generation. Scores like BLEU, ROUGE, METEOR, etc. are more suited for Machine Translation (MT) tasks \cite{BERTscore}. For example, they compare how similar sentence P is to translated-sentence \^{P}. Instead, we outline some metrics from the Coh-Metrix web tool that helps us further quantitatively evaluate the quality of text generated. With the goal of eliciting emotions, we claim that subjective analysis of generated poetry is superior to any available objective metrics.

\subsection{Coh-Metrix}

\begin{table*}
    \begin{center}
        \begin{tabular}{ | p{0.5in} | c | c | c | c | c | c | c | c | }
            \hline
            \textbf{Model} & \textbf{FRE} & \textbf{FKGL} & \textbf{IMGc} & \textbf{CNCc} & \textbf{LDTTRa} & \textbf{PCREFp} & \textbf{PCSYNp} & \textbf{PCNARp}\\
            \hline
            {anger} & 93.07 & 2.01 & 445.91 & 407.16 & 0.53 & 0.68 & 80.78 & 53.19 \\
            \hline
            {antici\-pation} & 100 & 0.83 & 440.93 & 403.10 & 0.40 & 7.78 & 83.65 & 81.86 \\
            \hline
            {joy} & 100 & 0.39 & 446.23 & 403.07 & 0.39 & 11.90 & 91.31 & 78.52 \\
            \hline
            {sadness} & 98.20 & 1.18 & 444.96 & 403.25 & 0.44 & 1.88 & 88.69 & 72.91 \\
            \hline
           {trust} & 100 & 0.16 & 434.66 & 412.72 & 0.33 & 18.14 & 84.61 & 91.31 \\
            \hline
            {dream} & 100 & 0 & 427.36 & 377.48 & 0.24 & 99.90 & 65.17 & 70.88 \\
            \hline
        \end{tabular}
    \end{center}
\caption{Average Coh-Metrix evaluations across 25 randomly selected poems from each model.}
\label{tab:cohmetrix}
\end{table*}

To provide a quantitative calculation of the caliber of text our models produce, we outline in this section relevant metrics from the University of Memphis Coh-Metrix tool \cite{Graesser2004CohMetrixAO}. Coh-Metrix is a text evaluation software kit and from it, we have chosen 8 forms of assessment. The first two, Flesch-Kincaid Grade Level (FKGL) and Flesch Reading Ease (FRE), are two standard measures that deal with text readability and ease \cite{klare1974assessing}. The FKGL scores a text from grade level 0 to 18, while the FRE score is a 0-100 index with 100 being an easily readable text. We aim to produce text that is readable by all, so a low FKGL score and high FRE score would be ideal.

The next metrics we employ evaluate at the word level. The word imageability (IMGc) and word concreteness (CNCc) scores measure content words on their ability to create an image in the reader's mind and their ability to appeal to a reader's senses, respectively \cite{mrcpyschodatabase}. We aim for our art to create a connection between the reader and poem, so we believe imageability and concreteness of content words are two good measures with this in mind. We also make use of three text easibility principal component scores: narrativity (PCNARp), referential cohesion (PCREFp), and syntactic simplicity (PCSYNp) \cite{Graesser2004CohMetrixAO}. The text easibility PC scores are percentile scales, and thus we aim for higher numbers for these scores. Finally, we make use of the Lexical Diversity Type:Token Ratio score (LDTTRa) for all words. LDTTRa measures the ratio of \textit{type} (unique) words to all \textit{tokens} in the text. Because our text is relatively short, we aim for a middle ground in the LDTTRa ratio, meaning there is uniqueness in the word choice of the text, but cohesion is still upheld.

Inspection of our Coh-Metrix results show that randomly selected poems from all models fall at or below the 2nd-grade reading level (in FKGL scores) and are greater than 93 on the FRE scale. This suggests generated poems are easily readable by the majority of viewers. Looking at the IMGc and CNCc scores, we see that our poems, except for the dream model concreteness, fall in the 400s. Words with higher imageability and concreteness fall around the low 600s while words that are lower fall around the upper 200s on this scale. These scores reveal that our models are generating text that is concrete in word choice and that paint a picture. Our dream model scoring lower in the concreteness is reasonable as the word choice of dreams tends to be more abstract. Lastly, percentile scores of PCSYNp and PCNARp show that the majority of models are producing poems that are both syntactically simplistic and narrative. Most PCREFp scores are on the lower end of the scale. We suspect the reason these scores are lower is because the poems are not necessarily related and were all input at once. Table \ref{tab:cohmetrix} highlights these scores for each poetry model.

\section{Conclusion \& Future Work}
In this paper we influenced automatic natural language generation to create poetry through the use of classified emotion poems and dream text. To do so, we first leveraged a word-level emotion lexicon to construct a meaning for emotion-eliciting text and used that text to train separate language models.  Next, we gathered data of dream records and employed transfer learning in attempts to generate dream-like poetry. The work reported in this paper seeks to create art in the form of auto-generated poetry while opening the door to more projects involving emotion-eliciting text-based tasks and influenced creative neural generation.

We would like to thank the reviewers for their feedback on this
project. Comments and suggestions from reviewers ---{} both those that the were
incorporated into this article and those on which we will report in future work ---{}
provide invaluable insight as to improving our results. Importantly, our continuing research
involves gathering a more comprehensive human-evaluation with a larger number of
reviewers and poems to be judged. We also wish to gather data for the
underrepresented emotion categories, leading, ideally, to a more robust language model for
each emotion. Our work thus far provides a baseline for introducing emotions into
generated text via a word-level lexicon, but we wish to employ
other tools ---{} segment-level lexicons, for example ---{} in an attempt to better capture the
contextual dependencies of emotion. Additionally, the word-level baseline we have
produced focuses on generating single-emotion text. We are interested in
examining poems of multiple emotions and different levels of intensity to expand
on this study. Finally, we wish to seek out additional forms of replicating
creativity that artists incorporate in their work.

\nocite{zhangpoetry}
\nocite{Deepspeare}
\nocite{attention}
\balance
\bibliographystyle{acl_natbib}
\bibliography{poetry}

\end{document}